\documentclass[11pt,a4paper]{article}
\usepackage[hyperref]{emnlp-ijcnlp-2019}
\usepackage{times}
\usepackage{latexsym}

\usepackage{url}

\usepackage{soul}
\usepackage{url}
\usepackage{graphicx}
\usepackage{amsmath}
\usepackage{booktabs}
\usepackage{color}
\usepackage{CJK}
\usepackage{enumitem}
\usepackage{subfigure}
\usepackage{algorithmic}
\usepackage{amsfonts,amssymb,bbm,epsfig,bm}
\usepackage{float}
\usepackage{multirow}

\usepackage{multirow}

\newcommand{\seq}[1]{\mathbf{#1}}

\aclfinalcopy % Uncomment this line for the final submission

\title{A Discrete CVAE for Response Generation on Short-Text Conversation}

\author{Jun Gao\textsuperscript{1}\thanks{\;\;Work done when Jun Gao was interning at Tencent AI Lab.} ,
Wei Bi\textsuperscript{2}\thanks{\;\;Corresponding author} ,
Xiaojiang Liu\textsuperscript{2},
Junhui Li\textsuperscript{1},
Guodong Zhou\textsuperscript{1},
Shuming Shi\textsuperscript{2}\\
\textsuperscript{1}{School of Computer Science and Technology, Soochow University, Suzhou, China}\\
imgaojun@gmail.com, \{lijunhui,gdzhou\}@suda.edu.cn\\
\textsuperscript{2}{Tencent AI Lab, Shenzhen, China}\\
{\{victoriabi, kieranliu, shumingshi\}@tencent.com}
}
\date{}

\begin{document}
\begin{CJK*}{UTF8}{gbsn}

\maketitle
\begin{abstract}
Neural conversation models such as encoder-decoder models are easy to generate bland and generic responses. Some researchers propose to use the conditional variational autoencoder (CVAE) which maximizes the lower bound on the conditional log-likelihood on a continuous latent variable. With different sampled latent variables, the model is expected to generate diverse responses.    
Although the CVAE-based models have shown tremendous potential, their improvement of generating high-quality responses is still unsatisfactory. In this paper, we introduce a discrete latent variable with an explicit semantic meaning to improve the CVAE on short-text conversation. 
A major advantage of our model is that we can exploit the semantic distance between the latent variables to maintain good diversity between the sampled latent variables. 
Accordingly, we propose a two-stage sampling approach to enable efficient diverse variable selection from a large latent space assumed in the short-text conversation task. Experimental results indicate that our model outperforms various kinds of generation models under both automatic and human evaluations and generates more diverse and informative responses.
\end{abstract}

\section{Introduction}
\noindent 
Open-domain response generation~~\cite{perez2011conversational,sordoni2015neural} for single-round short text conversation~\cite{shang2015neural}, aims at generating a meaningful and interesting response given a query from human users. 
Neural generation models are of growing interest in this topic due to their potential to leverage massive conversational datasets on the web.
These generation models such as encoder-decoder models~\cite{vinyals2015neural,shang2015neural,wen2015semantically}, directly build a mapping from the input query to its output response, which treats all query-response pairs uniformly and optimizes the maximum likelihood estimation (MLE). 
However, when the models converge, they tend to output bland and generic responses
~\cite{li2016diversity,li2016deep,serban2016building}.

Many enhanced encoder-decoder approaches have been proposed to improve the quality of generated responses.
They can be broadly classified into two categories (see Section~\ref{sec:related} for details):
(1) One that does not change the encoder-decoder framework itself.
These approaches only change the decoding strategy, such as encouraging diverse tokens to be selected in beam search~\cite{li2016diversity,li2016simple}; or adding more components based on the encoder-decoder framework, such as the Generative Adversarial Network (GAN)-based methods~\cite{xu2017neural,zhang2018generating,li2017adversarial} which
add discriminators to perform adversarial training;
(2) The second category modifies the encoder-decoder framework directly by
incorporating useful information as latent variables in order to generate more specific responses~\cite{yao2017towards,zhou2017mechanism}.
However, all these enhanced methods still optimize the MLE of the log-likelihood or the complete log-likelihood conditioned on their assumed latent information, and models estimated by the MLE naturally favor to output frequent patterns in training data.
%\textcolor{red}{
%Some other work integrates the Seq2seq as 
%and still assume to learn a direct mapping from the query to the response.
%}
%which employs the Seq2seq as the generator and adds a discriminator to
%
%Unlike the above models, 

Instead of optimizing the MLE, some researchers propose to use the conditional variational autoencoder (CVAE), which maximizes the lower bound on the conditional data log-likelihood on a continuous latent variable~\cite{zhao2017learning,shen2017conditional}.
Open-domain response generation is a one-to-many problem, in which a query can be associated with many valid responses.
The CVAE-based models
generally assume the latent variable follows a multivariate Gaussian distribution with a diagonal covariance matrix, which can capture the latent distribution over all valid responses.
%mainly have three network modules: the prior network to map a query to a latent variable; the posterior network to model the mapping from a pair of query and response to a latent variable; the generation network to generate a response given a query and a sampled latent variable. 
%CVAE consists of a prior network and a posterior network
With different sampled latent variables, the model is expected to decode diverse responses. 
%\textcolor{green}{Thus by introducing an intermediate latent variable, CVAE has potentials in solving the "one source, multiple targets" problems. 
Due to the advantage of the CVAE in modeling the response generation process, we focus on improving the performance of the CVAE-based response generation models.
%\textcolor{green}{The use of objective function in CVAE is more effective than that adopted in RL and GAN-based methods where the RL-based methods mainly modify the loss function as the reward to be optimized and the GAN-based methods employ the Seq2seq as the generator trained with a well-designed discriminator jointly}
%Although the CVAE-based models have shown great potential in generating high-quality responses, 
%their improvement of generating high-quality responses is still unsatisfactory. 
%As shown in in~\citeauthor{zhao2017learning}~\shortcite{zhao2017learning}, 

Although the CVAE has achieved impressive results on many generation problems~\cite{yan2016attribute2image,sohn2015learning}, recent results on response generation
show that the CVAE-based generation models still suffer from the low output diversity problem.
That is multiple sampled latent variables result in responses with similar semantic meanings.  
To address this problem,  extra guided signals are often used to improve the basic CVAE. \citeauthor{zhao2017learning}~\shortcite{zhao2017learning} use dialogue acts to capture the discourse variations in multi-round dialogues as guided knowledge. However, such discourse information can hardly be extracted for short-text conversation.

In our work, we 
propose a discrete CVAE (DCVAE), which utilizes a discrete latent variable with an explicit semantic meaning in the CVAE for short-text conversation. 
%Unlike previous enhanced Seq2seq models, we optimize the lower bound on the conditional data log-likelihood, forming the problem as a discrete conditional variational autoencoder (DCVAE).
Our model mitigates the low output diversity problem in the CVAE by exploiting the semantic distance between the latent variables to maintain good diversity between the sampled latent variables.
Accordingly, we propose a two-stage sampling approach to enable efficient selection of diverse variables from a large latent space assumed in the short-text conversation task.
%A two-stage sampling approach is further proposed for the variable sampling in the prior and posterior networks in the DCVAE to make use of the word similarity information. 
%the mapping from xxx to xx; and a generation network to model the query response similarity, and. KL divergence of the recognition network and prior network are used as the training loss. In the inference step, the prior network is used to first sample the cluster, then the keywords from the query. The final responses are generated given both the keywords and the context vector of the input. Our model could naturally generate multiple diverse responses if we sample multiple latent keywords in the inference.  
%
%We conduct an empirical study on a large benchmark dataset, and compare our model with several state-of-the-art response generation methods. Empirical results show that our model can generate more specific responses, and outperform existing methods under both automatic and human evaluations. We also provide analyses on the effectiveness of the discrete latent variables used in our model and the proposed two-stage sampling approach. All our codes and results will be publicly available.
%

To summarize, this work makes three contributions:
(1) We propose a response generation model for short-text conversation based on a DCVAE, which utilizes a discrete latent variable with an explicit semantic meaning and could generate high-quality responses. 
(2) A two-stage sampling approach is devised to enable efficient selection of diverse variables from a large latent space assumed in the short-text conversation task.
(3) Experimental results show that the proposed DCVAE with the two-stage sampling approach outperforms various kinds of generation models under both automatic and human evaluations, and generates more high-quality responses.
% All our codes and results will be publicly available.
All our code and datasets are available at \url{https://ai.tencent.com/ailab/nlp/dialogue}.
%\end{enumerate}

%\textcolor{red}{The remainder of the paper is organized as follows: After the related work in Section 2, we introduce our model in Section 3. Experimental results are shown in Section 4. In Section 5, we come to our conclusion and future work.}

\section{Related Work}
\label{sec:related}
In this section, we briefly review recent advancement
in encoder-decoder models and CVAE-based models for response generation.
%}
%This method can lead to the issue that decoders cannot be tailored to generate target sequences with specific properties of interest, such as cue words and topics.
%\textcolor{blue}{introduce two basic methods and state they optimized MLE and suffer from the generic response.}

\subsection{Encoder-decoder models}
Encoder-decoder models for short-text conversation 
~\cite{vinyals2015neural,shang2015neural}
maximize the likelihood of responses given queries. During testing, a decoder sequentially generates a response using search strategies such as beam search. However, these models frequently generate bland and generic responses.
%containing little information. %Researchers have begun to study how to generate diverse and informative responses.
%In the following, we briefly review different categories of approaches proposed to address this problem.

Some early work improves the quality of generated responses by modifying the decoding strategy.
For example,
 \citeauthor{li2016diversity}~\shortcite{li2016diversity} propose to use the maximum mutual information (MMI) to penalize general responses in beam search during testing. 
%Later, %\citet{li2017data} proposed a data distillation method to train a series of generation models using data with different levels of specificity and used a reinforcement learning model to choose the model best suited for decoding. 
Some later studies alter the data distributions according to different sample weighting schemes, encouraging the model to put more emphasis on learning samples with rare words~\cite{nakamura2018another,liu2018towards}. As can be seen, these methods focus on either pre-processing the dataset before training or post-processing the results in testing, with no change to encoder-decoder models themselves.

%\subsection{Enhanced Sequence-to-Sequence}
Some other work use encoder-decoder models as the basis and add more components to refine the response generation process.
 \citeauthor{xu2017neural}~\shortcite{xu2017neural} present a GAN-based model with an approximate embedding layer. \newcite{zhang2018generating}
 employ an adversarial learning method to directly optimize the lower bounder of the MMI objective~\cite{li2016diversity} in model training.
 These models employ the encoder-decoder models as the generator and focus on how to design the discriminator and optimize the generator and discriminator jointly.
%  leverage adversarial training that
% allows distributional matching of synthetic and real responses to foster response diversity and explicitly optimize a variational lower bound on pairwise mutual information between the query and the response to improve informativeness. 
Deep reinforcement learning is also applied to model future reward in chatbot after an encoder-decoder model converges~\cite{li2016deep,li2017adversarial}.
% Later, \citeauthor{li2017adversarial}~\shortcite{li2017adversarial} extends this method with more general rewards. 
%The RL-based methods mainly change the Seq2seq loss function to the designed reward as the optimized objective, while the GAN-based methods 
The above methods directly integrate the encoder-decoder models as one of their model modules and still do not actually modify the encoder-decoder models.

 %Another line of work is to change the Seq2seq directly.
 Many attentions have turned to incorporate useful information as latent variables in the encoder-decoder framework to improve the quality of generated responses.
 \citeauthor{yao2017towards}~\shortcite{yao2017towards} consider that a response is generated by a query and a pre-computed cue word jointly.
\citeauthor{zhou2017mechanism}~\shortcite{zhou2017mechanism} utilize a set of latent embeddings to model diverse responding mechanisms.
\citeauthor{xing2017topic}~\shortcite{xing2017topic} 
introduce pre-defined topics from an external corpus to augment the information used in response generation.
%\citeauthor{zhang2018learning}~\shortcite{zhang2018learning} present a controlled response generation mechanism to handle different utterance-response relationships in terms of specificity.
%\textcolor{red}{
%\citeauthor{yao2017towards}~\shortcite{yao2017towards} proposed to generate an informative by inferring a cue word estimated by PMI and
\citeauthor{gao2019generating}~\shortcite{gao2019generating} propose a model that infers latent words to generate multiple responses.
%\textcolor{green}{Although latent variable has been introduced, these models optimized with MLE objective still suffer from the vanishing latent variable problem~\cite{zhao2017learning} where the decoder tends to ignore the latent variable.}
These studies indicate that many factors in conversation are useful to model the variation of a generated response, but it is nontrivial to extract all of them.
Also, these methods
still optimize the MLE of the complete log-likelihood conditioned on their assumed latent information, and the model optimized with the MLE naturally favors to output frequent patterns in the training data.
Note that we apply a similar latent space assumption as used in \cite{yao2017towards,gao2019generating}, i.e. the latent variables are words from the vocabulary. However,  they use a latent word in a factorized encoder-decoder model, but our model uses it to construct a discrete CVAE and our optimization algorithm is entirely different from theirs. 

\subsection{The CVAE-based models}
%\textcolor{red}{
%Instead of optimizing the MLE,
A few works indicate that it is worth trying to apply the CVAE to dialogue generation which is originally used in image generation~\cite{yan2016attribute2image,sohn2015learning} and optimized with the variational lower bound of the conditional log-likelihood.
For task-oriented dialogues, 
\citeauthor{wen2017latent}~\shortcite{wen2017latent} use the latent variable to model intentions in the framework of neural variational inference.
For chit-chat multi-round conversations,
\citeauthor{serban2017hierarchical}~\shortcite{serban2017hierarchical} 
model the generative process with multiple levels of variability based on a hierarchical sequence-to-sequence model with a continuous high-dimensional latent variable.
\citeauthor{zhao2017learning}~\shortcite{zhao2017learning} make use of the CVAE and the latent variable is used to capture discourse-level variations. \citeauthor{gu2018dialogwae}~\shortcite{gu2018dialogwae} propose to 
%a latent variable model for multi-turn dialogues and 
induce the latent variables by transforming context-dependent Gaussian noise.
%Moreover, they propose a bag-of-words (BOW) loss and leverage knowledge-guided features to mitigate the vanishing KL distance problem.
\citeauthor{shen2017conditional}~\shortcite{shen2017conditional} present a conditional variational framework for generating specific responses based on specific attributes. 
%\citeauthor{du2018variational}~\shortcite{du2018variational} introduce a series of latent variables to condition the generation of each word in the response sequence.
Yet, it is observed in other tasks such as image captioning~\cite{wang2017diverse} and question generation~\cite{fan2018reinforcement} that the CVAE-based generation models suffer from the low output diversity problem, i.e. multiple sampled variables point to the same generated sequences.
%These CVAE conversations generally suffers from the mode collapse problem that multiple sampled latent variable point to the same generated result, resulting in poor output diversity. xxxx 
In this work, we utilize a discrete latent variable with an interpretable meaning to alleviate this low output diversity problem on short-text conversation.

We find that \citeauthor{zhao2018unsupervised}~\shortcite{zhao2018unsupervised} make use of a set of discrete variables that define high-level attributes of a response. 
Although they interpret meanings of the learned discrete latent variables by clustering data according to certain classes (e.g. dialog acts), such latent variables still have no exact meanings.
In our model, we connect each latent variable with a word in the vocabulary, thus each latent variable has an exact semantic meaning.
Besides, they focus on multi-turn dialogue generation and presented an unsupervised discrete sentence representation learning method learned from the context while our concentration is primarily on single-turn dialogue generation with no context information. 
%unsupervised latent discrete vector as xxxx

% We find that 
% %for task-oriented dialogues, 
% \citeauthor{wen2017latent}~\shortcite{wen2017latent} and \citeauthor{zhao2018unsupervised}~\shortcite{zhao2018unsupervised} independently propose to model the latent intentions/actions as discrete latent variables in the framework of neural variational inference.\textcolor{red}{
% \citeauthor{wen2017latent}~\shortcite{wen2017latent} target for the task-oriented dialogues, thus they use the intentions in the tasks as the latent variable; \citeauthor{zhao2018unsupervised}~\shortcite{zhao2018unsupervised} work on the multi-round dialogues, thus they apply the dialogue acts which captures discourse-level information as the latent variable. In our short text conversation setting, such signals are not applicable.
% These two methods are  similar to our model at first glance.}
% However, our model has two major differences.  First, we target for the chit-chat short text conversations specifically and utilize words in the vocabulary as the discrete latent variables to model the short-text-level variations. Second, in order to capture enough information in the latent space, our latent word space for the open-domain short text conversation task is assumed much larger than the intention/action space in the previous two work. To handle the large latent space, we further devise a two-stage sampling approach for training and inference.

%Discrete CVAE methods\footnote{***other work using discrete variables in neural dialog models and other dcvae papers}

\begin{figure*}[t]
    \centering
    \includegraphics[width=0.85\linewidth]{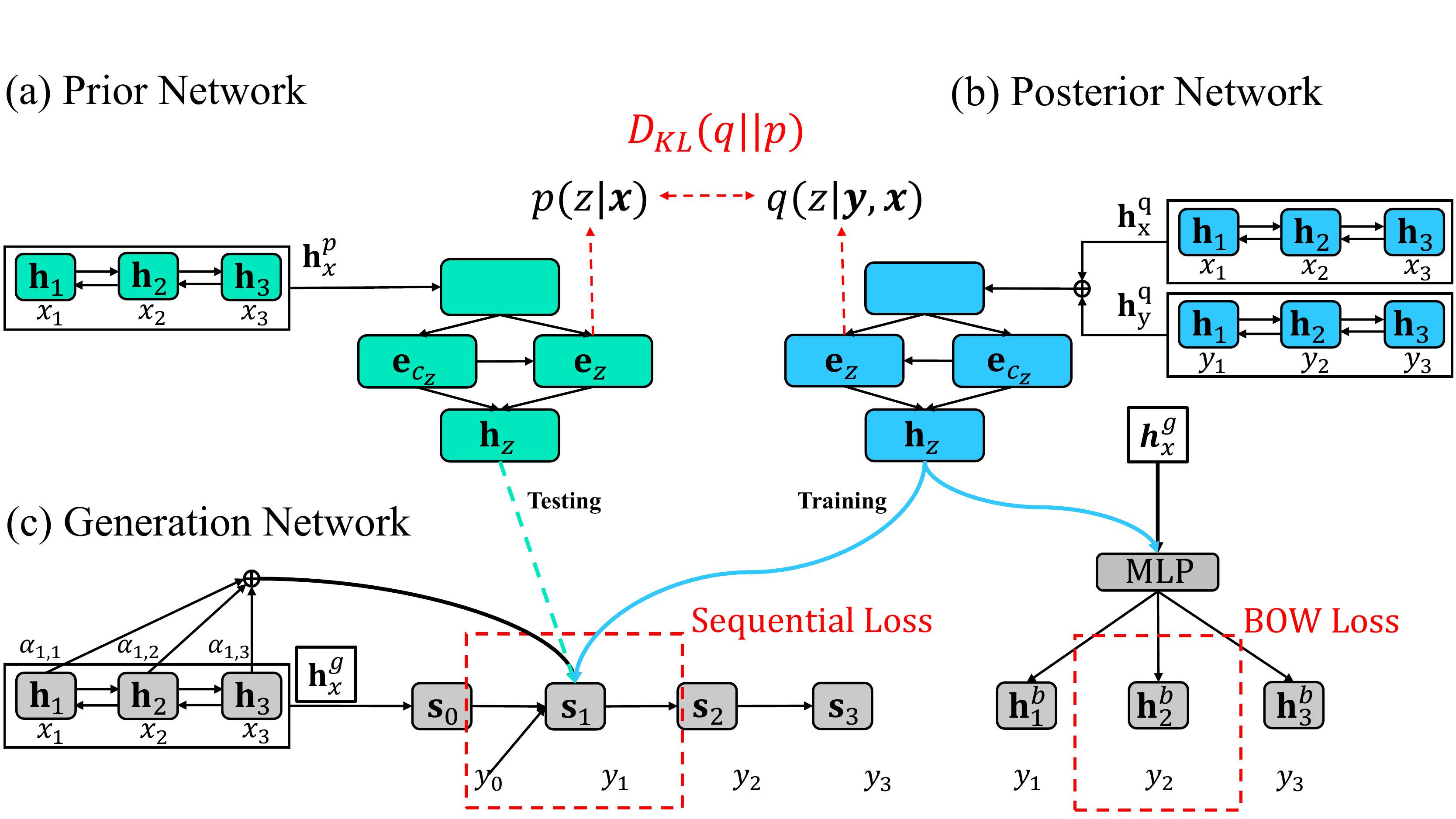}
    \caption{The architecture of the proposed discrete CVAE. $\seq{e}_{c_z}$ and $\seq{e}_z$ are embeddings of a cluster and a word sampled from the estimated discrete distributions. $\seq{e}_{c_z}$ is only applied when the two-stage sampling approach in Section~\ref{sec:two} is used. If $\seq{e}_{c_z}$ is applied, the latent representation $\seq{h}_z$ is  the sum of  $\seq{e}_{c_z}$ and $\seq{e}_z$; otherwise, $\seq{h}_z$ is $\seq{e}_z$. $\alpha$ denotes the attention weight. $\oplus$ denotes the sum of input vectors.
    %$h_{ctx}^p$ and $h_{ctx}^q$ are representation vectors of 
    }
    \label{fig:fig_model}
\end{figure*}

\section{Proposed Models}

\subsection{DCVAE and Basic Network Modules}
\label{sec:basic}
%The goal of a generative response generation model for short text conversation is usually to learn the conditional distribution $p(\seq{y} | \seq{x})$, where $\seq{x}$ is the input query sequence and $\seq{y}$  is the  target response sequence. 
Following previous CVAE-based generation models~\cite{zhao2017learning},
we introduce a latent variable $z$ for each input sequence
and our goal is to maximize the lower bound on the conditional data log-likelihood $p(\seq{y}|\seq{x})$, where $\seq{x}$ is the input query sequence and $\seq{y}$  is the  target response sequence:
%\footnote{***cite the original cvae paper}:
\begin{eqnarray}
\log p(\seq{y}|\seq{x})\!\! & \geq & \!\!\mathbb{E}_{z \sim q(z|\seq{y},\seq{x})}[\log p(\seq{y}|\seq{x},z)]\nonumber\\ 
&& \!\!\!\!- D_{KL}(q(z|\seq{y},\seq{x})||p(z|\seq{x})).
\end{eqnarray}
 Here, $p(z|\seq{x})$/$q(z|\seq{y},\seq{x})$/$p(\seq{y}|\seq{x},z)$ is parameterized by the prior/posterior/generation network respectively. $D_{KL}(q(z|\seq{y},\seq{x})||p(z|\seq{x}))$ is the Kullback-Leibler (KL) divergence between the posterior and prior distribution.
Generally, $z$ is set to follow a Gaussian distribution in both the prior and posterior networks.
%However, it is observed that in other tasks such as image captioning~\cite{wang2017diverse} and question generation~\cite{fan2018reinforcement,khullar2018automatic}, the resulting model using the above 
As mentioned in the related work, directly using the above CVAE formulation causes the low output diversity problem. This observation is also validated in the short-text conversation task in our experiments.

Now, we introduce our basic discrete CVAE formulation to alleviate the low output diversity problem.
%\begin{enumerate}[wide=0.5\parindent]
%\item 
We change the continuous latent variable $z$ to a discrete latent one with an explicit interpretable meaning, which could actively control the generation
of the response. 
An intuitive way is to connect each latent variable with a word in the vocabulary.
% \textcolor{blue}{By sampling different latent $z$'s, the generation network will make use of the word embeddings corresponding to these words as a part of its inputs for decoding.}
%In the full model introduced in Section~\ref{sec:two}, each latent variable $z$ is constructed by a word and its associated word cluster.
%There are mainly two advantages of our DCVAE:
%\begin{itemize}[wide=0.0\parindent]
    %\item Interpretability: 
    With a sampled latent $z$ from the prior (in testing)/posterior network (in training), the generation network will take the query representation together with the word embedding of this latent variable as the input to decode the response. 
Here, we assume that a single word is enough to drive the generation network to output diverse responses for short text conversation, in which the response is generally short and compact. 

    %\item 
    %Maintaining semantic distance in the latent space: 
    A major advantage of our DCVAE is that for words with far different meanings, their word embeddings (especially that we use a good pre-trained word embedding corpus) generally have a large distance and drive the generation network to decode scattered responses, thus improve the output diversity. In the standard CVAE, $z$'s assumed in a continuous space may not maintain the semantic distance as in the embedding space and diverse $z$'s may point to the same semantic meaning, in which case the generation network is hard to train well with such confusing information.  
    Moreover, we can make use of the semantic distance between latent variables to perform better sampling to approximate the objective during optimization, which will be introduced in Section~\ref{sec:two}.
%\end{itemize}
%This assumption is also made and validated in previous work~\cite{yao2017towards,gao2018generating}

%\textcolor{blue}{add more motivation and explanation: different z are with diverse meaning and their embedding are different, the generator will change a lot and the mode collapse is not that severe.}
%\item 

The latent variable $z$ is thus set to follow a categorical distribution with each dimension corresponding to a word in the vocabulary. Therefore the prior and posterior networks should output categorical probability distributions: 
\begin{eqnarray}
    p_{\theta}(z|\seq{x}) & = & \mbox{softmax}(g_{\theta}(\seq{x})), \label{prob_p} \label{eq:p}\\
    q_{\phi}(z|\seq{y},\seq{x}) &= &\mbox{softmax}(f_{\phi}(\seq{y},\seq{x})), \label{prob_q}
\end{eqnarray}
where $\theta$ and $\phi$ are parameters of the two networks respectively.
The KL distance of these two distributions can be calculated in a closed form solution:
\begin{eqnarray}
%\begin{split}
\lefteqn{D_{KL}(q(z|\seq{y},\seq{x})||p(z|\seq{x})) = }  \nonumber\\
&\quad\quad\quad\sum_{z \in Z} q(z|\seq{y},\seq{x})\log \frac{q(z|\seq{y},\seq{x})}{p(z|\seq{x})},
%\end{split}
\end{eqnarray}
where $Z$ contains all words in the vocabulary. 
%Here we set it as the vocabulary used for training.
%\end{enumerate}
\noindent
In the following, we present the details of the prior, posterior and generation network.

\noindent \textbf{Prior network $p(z|\seq{x})$}:
%The prior network
It aims at inferring the latent variable $z$ given the input sequence $x$. 
 We first obtain an input representation $\seq{h}_{\seq{x}}^{p}$ by encoding the input query $\seq{x}$ with a bi-directional GRU 
  and then compute $g_{\theta}(\seq{x})$ in Eq.~\ref{eq:p}  as follows:
\begin{eqnarray}
 g_{\theta}(\seq{x}) = {\seq{W}_2}
  \cdot \mbox{tanh}( \seq{W}_1  \seq{h}_{x}^{p} + \seq{b}_1 ) +  \seq{b}_2,    
  \label{eq:prior} 
\end{eqnarray}
where $\theta$ contains parameters in both the bidirectional GRU and Eq.~\ref{eq:prior}. 

\noindent \textbf{Posterior network $q(z|\seq{y}, \seq{x})$}:
It infers a latent variable $z$ given a input query $\seq{x}$ and its target response $\seq{y}$. We construct both representations for the input and the target sequence by separated bi-directional GRU's, then add them up to compute $f_{\phi}(\seq{y}, \seq{x})$ in Eq.~\ref{prob_q} to predict the probability of $z$:
\begin{eqnarray}
 \!f_{\phi}(\seq{y}, \seq{x}) \! = \!
  \seq{W}_4 
 \! \cdot \! \mbox{tanh}( \seq{W}_3  (\seq{h}_{x}^{q}\! +\! \seq{h}_{y}^q) 
  \!+\! \seq{b}_3 )\! + \!   \seq{b}_4, \!\!\!
 \label{eq:post}
 \end{eqnarray}
where $\phi$ contains parameters in the two encoding functions and Eq.~\ref{eq:post}. Note that the parameters of the encoding functions are not shared in the prior and posterior network.

%represent the dialog context. Finally it takes a form similar to the prior network, which first sample the cluster based on $q(z|\seq{y},\seq{x})$, next sample a specific $z$ from $z$ within the sampled cluster based on $q(z|\seq{x},\seq{y}, c_{k_z})$.

\noindent \textbf{Generation network} $p(\seq{y}|\seq{x},z)$:
We adopt an encoder-decoder model with attention~\cite{luong2015effective} used in the decoder.
With a sampled latent variable $z$, a typical strategy is to combine its representation, which in this case is the word embedding $\seq{e}_z$ of $z$, only in the beginning of decoding. 
However, many previous works observe that the influence of the added information will vanish over time~\cite{yao2017towards,gao2019generating}.
Thus, 
after obtaining an attentional hidden state at each decoding step,
we concatenate the representation $\seq{h}_z$ of the latent variable and the current hidden state to produce a final output in our generation network.

%%%%%%%%%%%%%%%%%%%%%%%%%%%%%%%%%%%%%%
\subsection{A Two-Stage Sampling Approach}
\label{sec:two}
%\textcolor{blue}{how the training with your objective can be performed?}
When the CVAE models are optimized, they tend to converge to a solution with a vanishingly
small KL term,  thus failing to encode meaningful information in $z$.
To address this problem, we follow the idea in~\cite{zhao2017learning}, which introduces an auxiliary loss that requires the decoder in the generation network
to predict the bag-of-words in the response $\seq{y}$. 
Specifically,  the response $\seq{y}$ is now represented by two sequences simultaneously: $\seq{y}_o$ with word order and $\seq{y}_{bow}$ without order. These two sequences are assumed to be conditionally independent given $z$ and $\seq{x}$.
Then our training objective can be rewritten as:
\begin{equation}
    \begin{split}
        J(\Theta) = &\mathbb{E}_{z \sim q(z|\seq{y},\seq{x})}[\log p(\seq{y}|\seq{x},z)]  \\
        &- D_{KL}(q(z|\seq{y},\seq{x})||p(z|\seq{x})) \\
        &+ \mathbb{E}_{z \sim q(z|\seq{y},\seq{x})}[\log p(\seq{y}_{bow}|\seq{x}, z)],
    \end{split}
\end{equation}
where $p(\seq{y}_{bow}|\seq{x}, z)$ is obtained by a multilayer perceptron $\seq{h}^{b} = \mbox{MLP}(\seq{x}, z)$:
\begin{equation}
 p(\seq{y}_{bow}|\seq{x}, z) =  \prod_{t=1}^{|\seq{y}|}\frac{\exp(h_{y_t}^{b})}{\sum_{j \in V}\exp(h_{j}^{b})},
\end{equation}
where $|\seq{y}|$ is the length of $\seq{y}$, $y_t$ is the word index of $t$-{th} word in $\seq{y}$, and $V$ is the vocabulary size. 

%We perform two samplings in our model. One is that 
During training, we generally approximate $\mathbb{E}_{z \sim q(z|\seq{y},\seq{x})}[\log p(\seq{y}|\seq{x},z)]$ by sampling $N$ times of $z$ from the distribution $q(z|\seq{y}, \seq{x})$.
%The other is that in testing, we sample $z$ from the prior $p(z|\seq{x})$ and integrate it in the generation network for response generation. 
%
%However, we often requires a large space of $z$. xxxxx
In our model, the latent space is discrete but generally large since we set it as the vocabulary in the dataset
\footnote{Note that we remove special tokens including UNK (unknown token), BOS (start of sentence) and EOS (end of sentence) in the latent space such that our model will only select meaningful words as the latent variables.}.
%Also, for a given short-text input $\seq{x}$, only very few $z$'s will provide useful information to guide the response generation since most words in the vocabulary are irrelevant to the content of $\seq{x}$. 
The vocabulary consists of words that are similar in syntactic or semantic.
Directly sampling $z$ from the categorical distribution in Eq.~\ref{prob_q} cannot make use of such word similarity information.

Hence, we propose to modify our model in Section~\ref{sec:basic} to consider the word similarity for sampling multiple accurate and diverse latent $z$'s. We first cluster $z \in Z$ into $K$ clusters $c_1,\ldots, c_K$. Each $z$ belongs to only one of the $K$ clusters and dissimilar words lie in distinctive groups. We use the K-means clustering algorithm to group $z$'s using a pre-trained embedding corpus~\cite{song2018directional}. 
%should consider the common characteristics and ignore the specific characteristics among the individual features to generalize the specific features.
%
%\textcolor{blue}{the motivation of two-step is not clear. refer to the nips paper why we need to perform clustering.}
%
Then we revise the  posterior network to perform a two-stage cluster sampling by decomposing $q(z|\seq{y}, \seq{x})$ as :
\begin{eqnarray}
q(z|\seq{y}, \seq{x}) &=& \sum_{k}q(z|\seq{x}, \seq{y}, c_k) q(c_k|\seq{x}, \seq{y})\nonumber\\
&=&  q(z|\seq{x}, \seq{y},c_{k_z})q(c_{k_z}|\seq{x}, \seq{y}).
%    q(z|\seq{y},\seq{x}) &= & \sum_{k}q(z|\seq{x}, \seq{y}, c_k) q(c_k|\seq{x}) = q(z|\seq{x}, \seq{y}, \seq{y}, c_{k_z})q(c_{k_z}|\seq{x}, 
%\seq{y})
\end{eqnarray}
That is, we first compute $q(c_{k_z}|\seq{y}, \seq{x})$, which is the probability of the cluster that $z$ belongs to conditioned on both $\seq{x}$ and $\seq{y}$. Next, we compute $q(z|\seq{x}, \seq{y}, c_{k_z})$, which is the probability distribution of $z$ conditioned on the $\seq{x}$, $\seq{y}$ and the cluster $c_{k_z}$.
When we perform sampling from $q(z|\seq{x}, \seq{y})$, we can exploit the following two-stage sampling approach: first sample the cluster based on $q( c_{k} |\seq{x}, \seq{y})$; next sample a specific $z$ from $z$'s within the sampled cluster based on $q(z|\seq{x}, \seq{y}, c_{k_z})$.

Similarly, we can decompose the prior distribution $p(z| \seq{x})$ accordingly for consistency: 
\begin{eqnarray}
p(z|\seq{x}) = p(z|\seq{x}, c_{k_z})p(c_{k_z}|\seq{x}).
\end{eqnarray}
In testing, we can perform the two-stage sampling according to 
$p(c_{k}|\seq{x})$ and $p(z|\seq{x}, c_{k_z})$. Our full model is illustrated in Figure~\ref{fig:fig_model}.

\noindent 
\textbf{Network structure modification:}
To modify the network structure for the two-stage sampling method, we first compute the probability of each cluster given $\seq{x}$ in the prior network (or $\seq{x}$ and $\seq{y}$ in the posterior network) with a softmax layer (Eq.~\ref{eq:prior} or Eq.~\ref{eq:post} followed by a softmax function). We then add the input representation and the cluster embedding $\seq{e}_{c_z}$ of a sampled cluster $c_{z}$, and use another softmax layer to compute the probability of each $z$ within the sampled cluster. 
In the generation network, the representation of $z$ is the sum of the cluster embedding $\seq{e}_{c_z}$ and its word embedding $\seq{e}_{z}$.
%Note that in our network training, we share the same set of word embeddings for all network modules, but other parameters such as those in the encoders are different for each module.
%The posterior network can be modified in a similar way.

\noindent 
\textbf{Network pre-training:}
To speed up the convergence of our model, we pre-extract keywords from each query using the TF-IDF method. Then we use these keywords to pre-train the prior and posterior networks. 
The generation network is not pre-trained because in practice it converges fast in only a few epochs.

%%%%%%%%%%%%%%%%%%%%%%%%%%%%%%%%%%%%%%%%%%%%%%%%%%%%%%%%%%%%%%%%%%%%%%%%%%%%%%%%%%%%%%%%%%%%%%%%%%%%%%%%%%%%

\section{Experimental Settings}
Next, we describe our experimental settings including the dataset, implementation details, all compared methods, and the evaluation metrics. %All our codes and results will be publicly available.

\subsection{Dataset}
\label{sec:dataset}
We conduct our experiments on a short-text conversation benchmark dataset~\cite{shang2015neural} which contains about 4 million post-response pairs from the Sina Weibo~\footnote{\url{http://weibo.com}}, a Chinese social platforms. We employ the Jieba Chinese word segmenter~\footnote{\url{https://github.com/fxsjy/jieba}} to tokenize the queries and responses into sequences of Chinese words. 
We use a vocabulary of 50,000 words (a mixture of Chinese words and characters), which covers 99.98\% of words in the dataset. All other words are replaced with $<$UNK$>$.
% We randomly hold out two subsets as the development and test dataset, each containing 900 pairs.

\subsection{Implementation Details}

We use single-layer
bi-directional GRU for the encoder in the prior/posterior/generation network, and one-layer GRU for the decoder in the generation network. The dimension of all hidden vectors is 1024. The cluster embedding dimension is 620. Except that the word embeddings are initialized by the word embedding corpus~\cite{song2018directional}, all other parameters are initialized by sampling from a uniform distribution $[-0.1,0.1]$. The batch size is 128. We use Adam optimizer with a learning rate of 0.0001. 
For the number of clusters $K$ in our method, we evaluate four different values $(5, 10, 100, 1000)$ using automatic metrics and  set $K$ to 10 which tops the four options empirically.
It takes about one day for every two epochs of our model on a Tesla P40 GPU, and we train ten epochs in total. During testing, we use beam search with a beam size of 10. 

\renewcommand\arraystretch{1.1}
\begin{table*}[tb]
    \small
    %\vspace{0em}
    \begin{center}
        \setlength{\tabcolsep}{1.5mm}
        \begin{tabular}{ l | c | c | c | c | c | c || c|c|c}
            \hline
            %    \cmidrule[\heavyrulewidth]{1-4}
            \multirow{2}{*}{Method} & \multicolumn{6}{c}{Automatic Metrics} & \multicolumn{3}{c}{Human Evaluation}\\
            & BLEU-1 & BLEU-2 & BLEU-3 & BLEU-4 &dist-1 &dist-2 & quality & accept & good \\
            \hline
            Seq2seq & $0.63 \pm 0.14$& $0.49 \pm 0.16$& $0.35 \pm 0.14$& $0.21 \pm 0.12$ &$0.03$&$0.08$ & $1.64 \pm 0.30$  & $49\%$& $15\%$\\
        %    DBS &0.472 & 0.343& 0.238& $0.147$ &$0.037$&$0.103$\\
            MMI-bidi & $0.54 \pm 0.17$& $0.39 \pm 0.18$& $0.28 \pm 0.15$& $0.17 \pm 0.13$ &$0.04$&$0.11$ & $1.71 \pm 0.34$ & $52\%$& 19\%\\
            %  \hline
            CVAE & $0.60 \pm 0.13$& $0.43 \pm 0.16$& $0.30 \pm 0.14$& $0.18 \pm 0.11$ &$0.03$ &$0.06$ & $1.60 \pm 0.33$& $47\%$& 13\%\\
            MANM & $0.62 \pm 0.14$& $0.48 \pm 0.15$& $0.34 \pm 0.14$& $0.22 \pm 0.12$ &$0.05$ &$0.14$ & $1.73 \pm0.35$ & $53\%$& 21\%\\
            HGFU & $0.52 \pm 0.11$& $0.38 \pm 0.14$& $0.27 \pm 0.12$& $0.16 \pm 0.11$ &$0.08$ &$0.27$ & $1.63 \pm 0.39$& $51\%$& $12\%$\\
            \hline
            DCVAE & $\mathbf{0.64 \pm 0.14}$& $\mathbf{0.49 \pm 0.16}$& $\mathbf{0.35 \pm 0.15}$& $\mathbf{0.22 \pm 0.13}$ &$0.08$&$0.24$ & $\mathbf{2.03 \pm 0.34}$ & \textbf{73\%}& \textbf{30\%}\\
            \hline
            %    \bottomrule
        \end{tabular}
%         \vspace{-5pt}
        \caption{The automatic and human evaluation results of all compared methods. Note that the acceptable ratio is the percentage of responses with 2 or 3 points.}
        \label{tab:weibo_auto}
%         \vspace{-10pt}
    \end{center}
\end{table*}

\subsection{Compared Methods}
In our work, we focus on comparing various methods that model $p(\seq{y}|\seq{x})$ differently. We compare our proposed discrete CVAE ({\bf DCVAE}) with the two-stage sampling approach to three categories of response generation models:
\begin{enumerate}[wide=0.5\parindent]
\item Baselines: {\bf Seq2seq}, the basic encoder-decoder model with
soft attention mechanism ~\cite{bahdanau2015neural} used in decoding and  beam search used in testing;
{\bf MMI-bidi}~\cite{li2016diversity}, which uses the MMI to re-rank results from beam search.
\item {\bf CVAE}~\cite{zhao2017learning}: We adjust the original work which is for multi-round conversation for our single-round setting.
For a fair comparison, we utilize the same keywords used in our network pre-training as the knowledge-guided features in this model.
\item Other enhanced encoder-decoder models:
 Hierarchical Gated Fusion Unit ({\bf HGFU})~\cite{yao2017towards}, which incorporates a cue word extracted using pointwise mutual information (PMI) into the decoder to generate meaningful responses;
 Mechanism-Aware Neural Machine ({\bf MANM})~\cite{zhou2017mechanism}, which introduces latent embeddings to allow for multiple diverse response generation.
\end{enumerate}
\noindent
Here, we do not compare RL/GAN-based methods because all our compared methods can replace their objectives with the use of reward functions in the RL-based methods or add a discriminator in the GAN-based methods to further improve the overall performance.
However, these are not the contribution of our work, which we leave to future work to discuss the usefulness of our model as well as other enhanced generation models combined with the RL/GAN-based methods.

\subsection{Evaluation}
\label{sec:evaluation}
%Evaluating the response generation models for open-domain short-text conversation is a very challenging problem. to employ both automatic and human evaluations.
To evaluate the responses generated by all compared methods,
we compute the following automatic metrics on our test set:
%\textcolor{red}{
\begin{enumerate}[wide=0\parindent]
    \item BLEU: 
    %It has been proved strongly correlated with human evaluations. 
    BLEU-n measures the average n-gram precision on a set of reference responses. We report BLEU-n with n=1,2,3,4.
    \item Distinct-1 \& distinct-2 ~\cite{li2016diversity}: We count the numbers of distinct uni-grams and bi-grams in the generated responses and divide the numbers by the total number of generated uni-grams and bi-grams in the test set. These metrics can be regarded as an automatic metric to evaluate the diversity of the responses.
    \end{enumerate}
Three annotators from a commercial annotation company are recruited to conduct our human evaluation.
Responses from different models are shuffled for labeling. 300 test queries are randomly selected out, and annotators are asked to independently score the results of these queries with different points in terms of their quality: (1) Good (3 points): The response is grammatical, semantically relevant to the query, and more importantly informative and interesting; (2) Acceptable (2 points): The response is grammatical, semantically relevant to the query, but  too trivial or generic (e.g.,``我不知道(I don't know)", ``我也是(Me too)'', ``我喜欢(I like it)" etc.); (3) Failed (1 point): The response has grammar mistakes or irrelevant to the query.
%     \begin{enumerate}[wide=0\parindent]
%     \item Good (3 points): The response is grammatical, semantically relevant to the query, and more importantly informative and interesting;
%     \item Acceptable (2 points): The response is grammatical, semantically relevant to the query, but  too trivial or generic (e.g.,``我不知道(I don't know)", ``我也是(Me too)'', ``我喜欢(I like it)" etc.);
%     \item Failed (1 point): The response has grammar mistakes or irrelevant to the query. 
% \end{enumerate}

\section{Experimental Results and Analysis}
In the following, we will present results of all compared methods and conduct a case study on such results. Then, we will perform further analysis of our proposed method by varying different settings of the components designed in our model.

%\section{Results and Analysis}

\subsection{Results on All Compared Methods}
\label{sec:result1}
Results on automatic metrics are shown on the left-hand side of Table~\ref{tab:weibo_auto}. 
%We first observe that compared with the Seq2seq baseline, MMI-bidi improves the distinct-1 and distinct-2 ratios but have worse BLEU scores; CVAE  has poor performance on both BLEU scores and distinct ratios. HGFU achieves the best distinct-1 and distinct-2 ratios among all models. However, it does not show any advantage of the BLEU scores. 
From the results we can see that our proposed DCVAE achieves the best BLEU scores and the second best distinct ratios. The HGFU has the best dist-2 ratio, but its BLEU scores are the worst. These results indicate that the responses generated by the HGFU are less close to the ground true references.
Although the automatic evaluation generally indicates the quality of generated responses, it can not accurately evaluate the generated response and the automatic metrics may not be consistent with human perceptions~\cite{liu2016not}. Thus, we consider human evaluation results more reliable. 

%Thus, we consider DCVAE performs the best considering all automatic metrics.
%
%Ours yields the best BLEU-1/2/3/4 scores and the second best distinct ratios among all methods, which are very close to the best ones by HGFU. Considering both accuracy and diversity, Ours performs the best.
%
For the human evaluation results on the right-hand side of Table~\ref{tab:weibo_auto}, we show the mean and standard deviation of all test results as well as the percentage of acceptable responses (2 or 3 points) and good responses (3 points only). 
Our proposed DCVAE has the best quality score among all compared methods. Moreover, DCVAE achieves a much higher good ratio, which means it generates more informative and interesting responses. 
Besides, the HGFU's acceptable and good ratios are much lower than our model indicating that it may not maintain enough response relevance when encouraging diversity. This is consistent with the results of the automatic evaluation in Table~\ref{tab:weibo_auto}.
We also notice that the CVAE achieves the worst human annotation score. 
This validates that the original CVAE for open-domain response generation does not work well and our proposed DCVAE is an effective way to improve the CVAE for better output diversity.

\begin{figure*}[t]
    \centering
    \includegraphics[width=\linewidth]{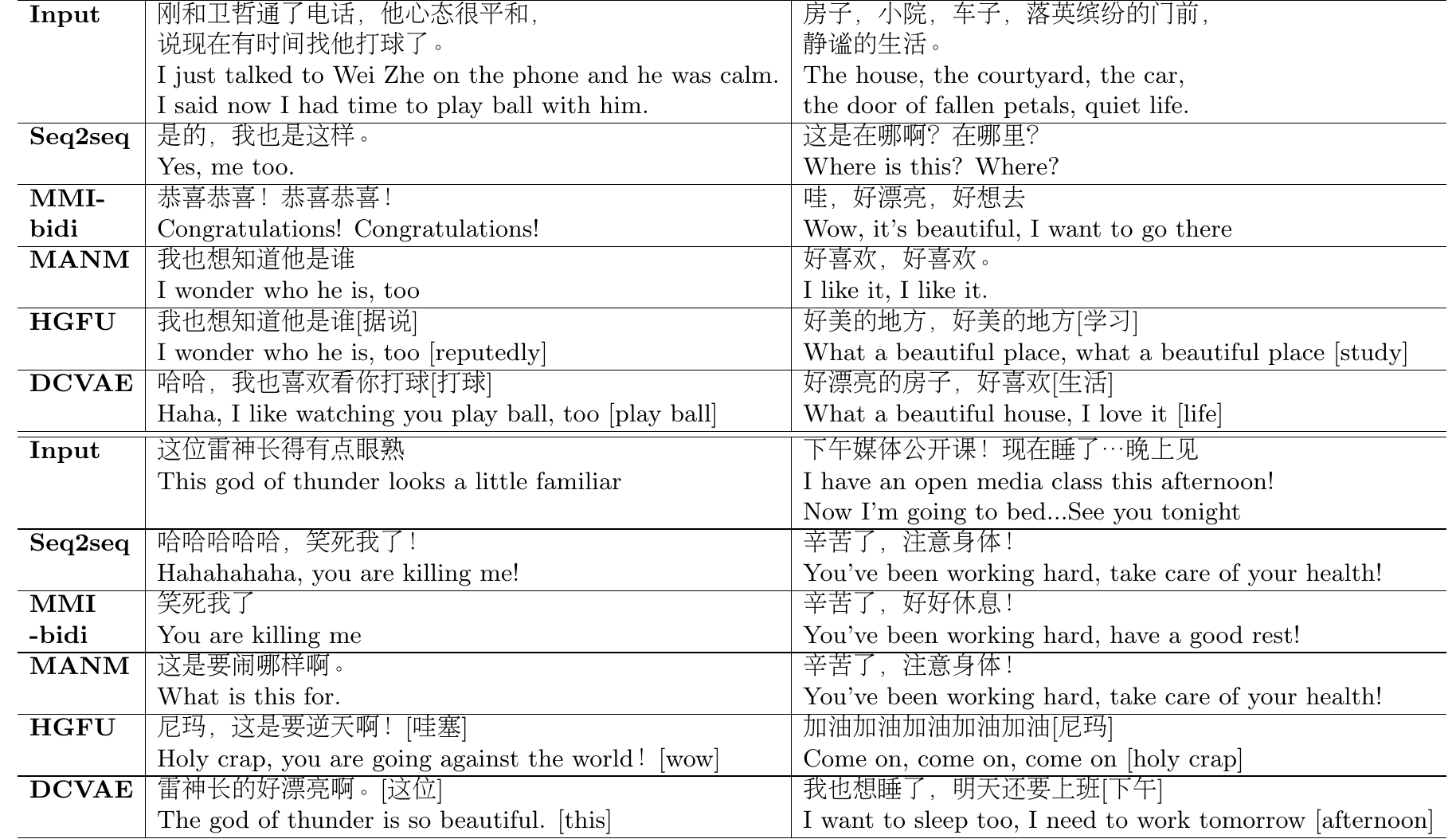}
%     \vspace{-5}
    \caption{Examples of the generated responses. The sampled latent words ($z$) are showed in the brackets.}
    \label{fig:fig_case}
%     \vspace{-10}
\end{figure*}

\subsection{Case Study}
\label{sec:case}
%During testing, our proposed model first samples a latent variable (i.e. a word) and then generates a response conditioned on the latent variable and the input query. 
Figure~\ref{fig:fig_case} shows four example queries with their responses generated by all compared methods. 
%\footnote{***how to select the case. check how others write this.}
%As can be seen, 
The Seq2seq baseline tends to generate less informative responses. Though MMI-bidi can select different words to be used, its generated responses are still far from informative.
MANM can avoid generating generic responses in most cases, but sometimes its generated response is irrelevant to the query, as shown in the left bottom case. Moreover, the latent responding mechanisms in MANM have no explicit or interpretable meaning. 
Similar results can be observed from HGFU. 
%TThis is also consistent with our analysis in the introduction that 
If the PMI selects irrelevant cue words, the resulting response may not be relevant.
Meanwhile, responses generated by our DCVAE are more informative as well as relevant to input queries.

\subsection{Different Sizes of the Latent Space}

\begin{figure}[tbp]
\begin{center}
\subfigure{
 \includegraphics[width=0.45\linewidth]{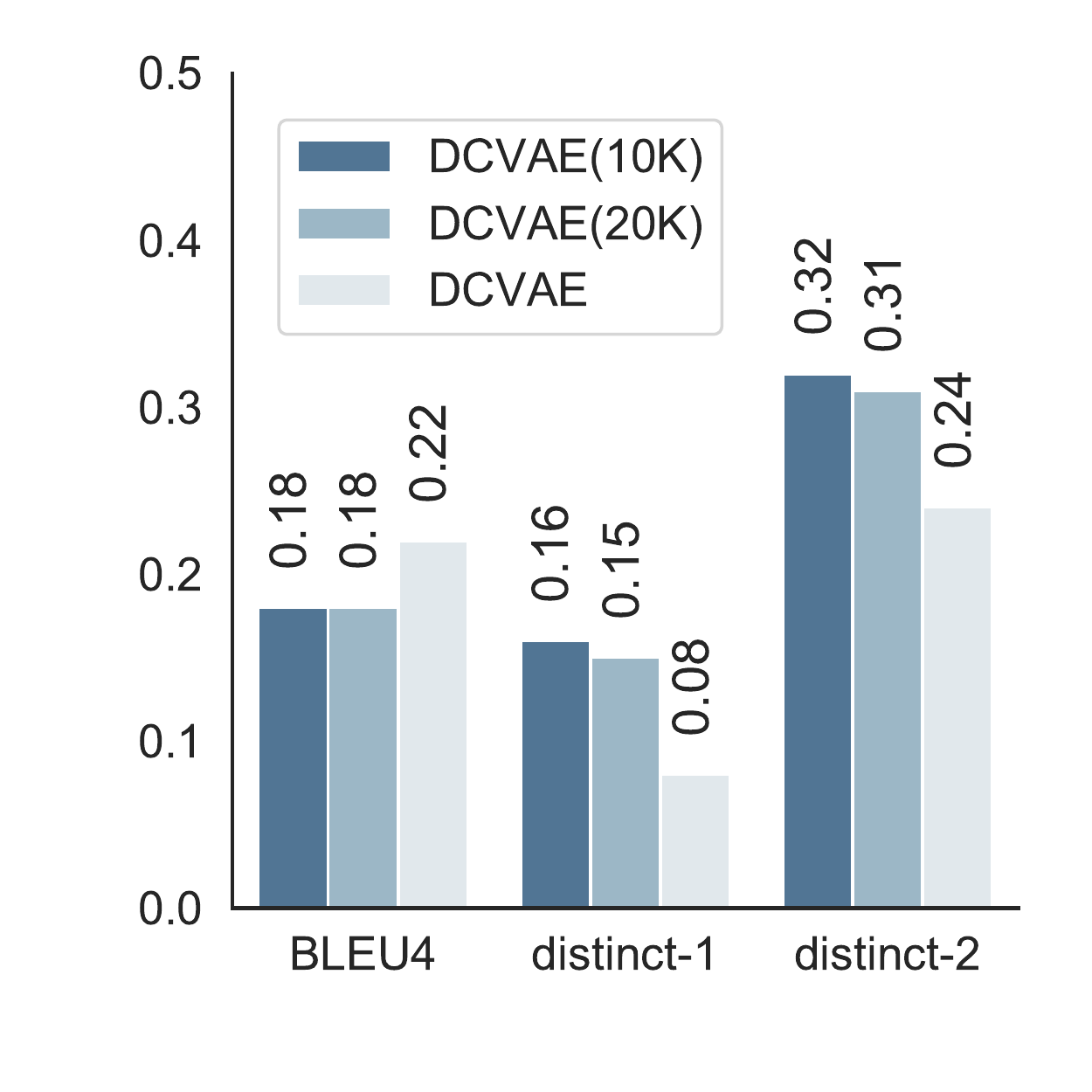}
 }
 \subfigure{
  \includegraphics[width=0.45\linewidth]{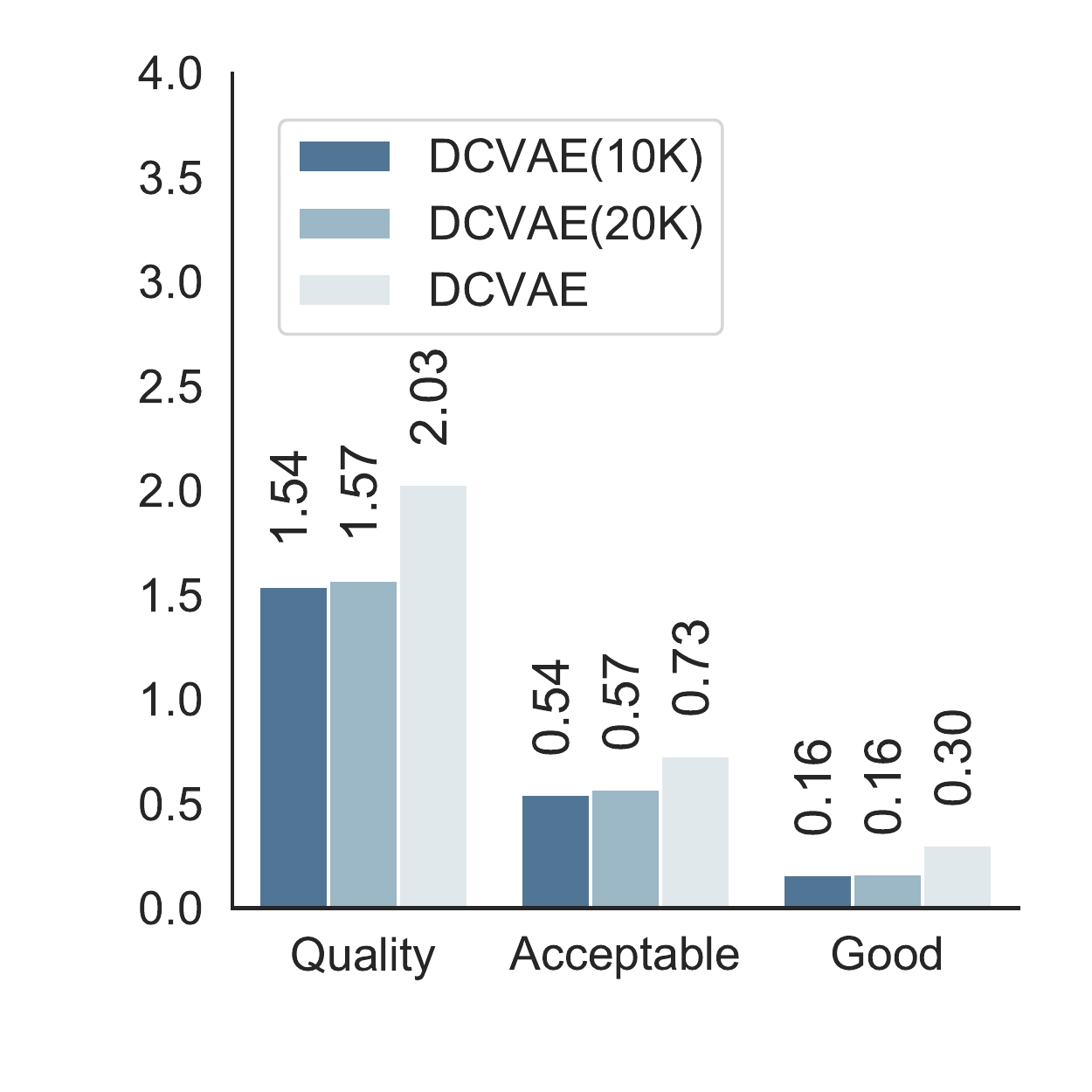}
 }
\end{center}
% \vspace{-5pt}

\caption{
Different sizes of the latent space used in the DCVAE: automatic evaluation (left) and
human evaluation (right).}
\label{fig:latent_word}
%\label{fig:attention}
% \vspace{-10pt}
\end{figure}

\noindent
We vary the size of the latent space (i.e., sampled word space $Z$) used in our proposed DCVAE. 
Figure~\ref{fig:latent_word} shows the automatic and human evaluation results on the latent space setting to the top 10k, 20k, all words in the vocabulary.  
On the automatic evaluation results, if the sampled latent space is getting larger, the BLEU-4 score increases but the distinct ratios drop. 
We find out that though the DCVAE with a small latent space has a higher distinct-1/2 ratio, many generated sentences are grammatically incorrect. This is also why the BLEU-4 score decreases.
On the human evaluation results, all metrics improve with the use of a larger latent space. 
This is consistent with our motivation that open-domain short-text conversation covers a wide range of topics and areas, and the top frequent words are not enough to capture the content of most training pairs. Thus a small latent space, i.e. the top frequent words only, is not feasible to model enough latent information and a large latent space is generally favored in our proposed model.
%\footnote{***it would be great if authors can conduct more analysis to find the reasons.
%What's the differences in terms of the predicted $z$ given different latent size. Is it because the top freq words mostly stop words and topics words are usually rare?}

\subsection{Analysis on the Two-Stage Sampling}

\begin{figure*}[tb]
\begin{center}
\subfigure[]{\label{fig:stage_auto}
 \includegraphics[width=0.3\linewidth]{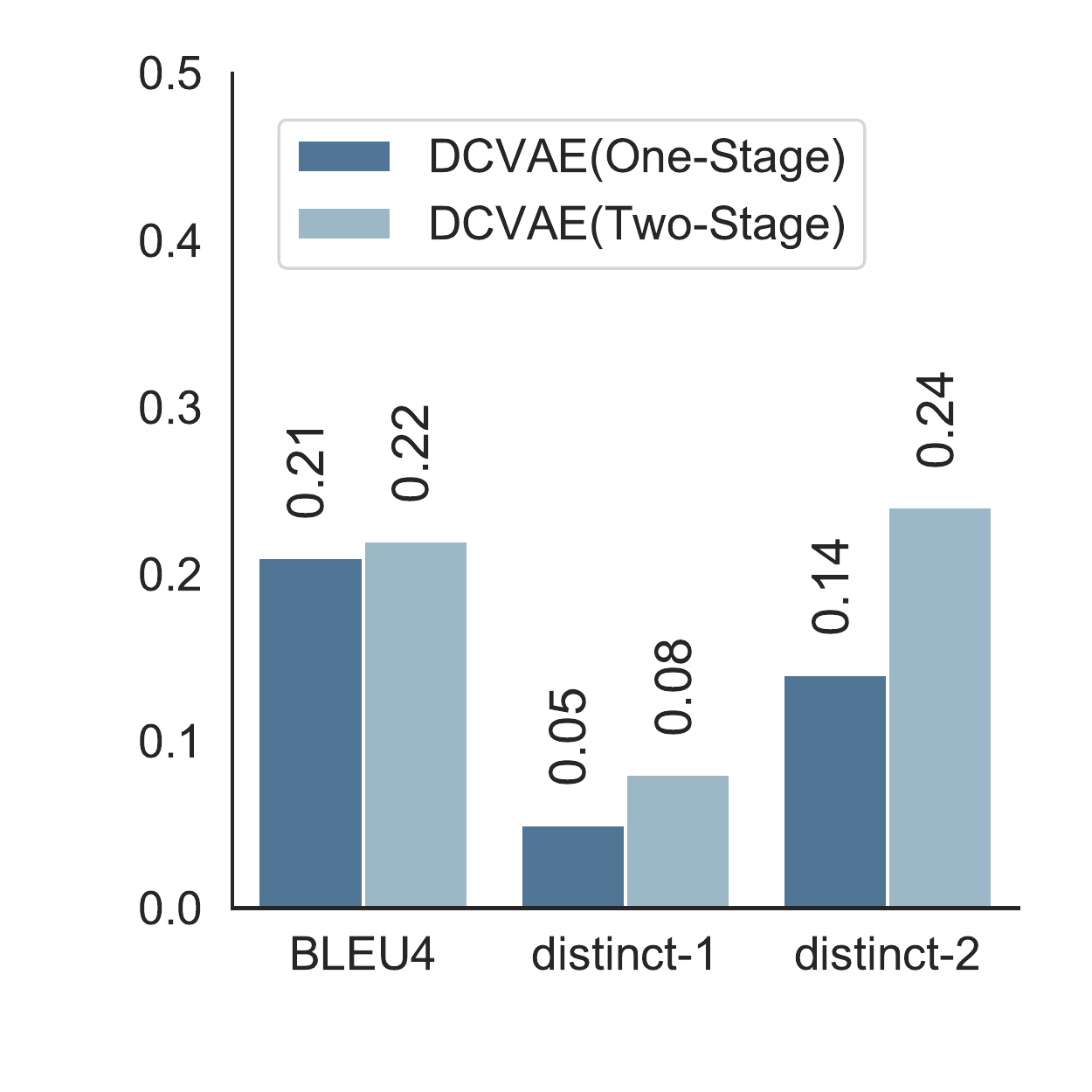}}
 \subfigure[]{\label{fig:stage_human}
  \includegraphics[width=0.3\linewidth]{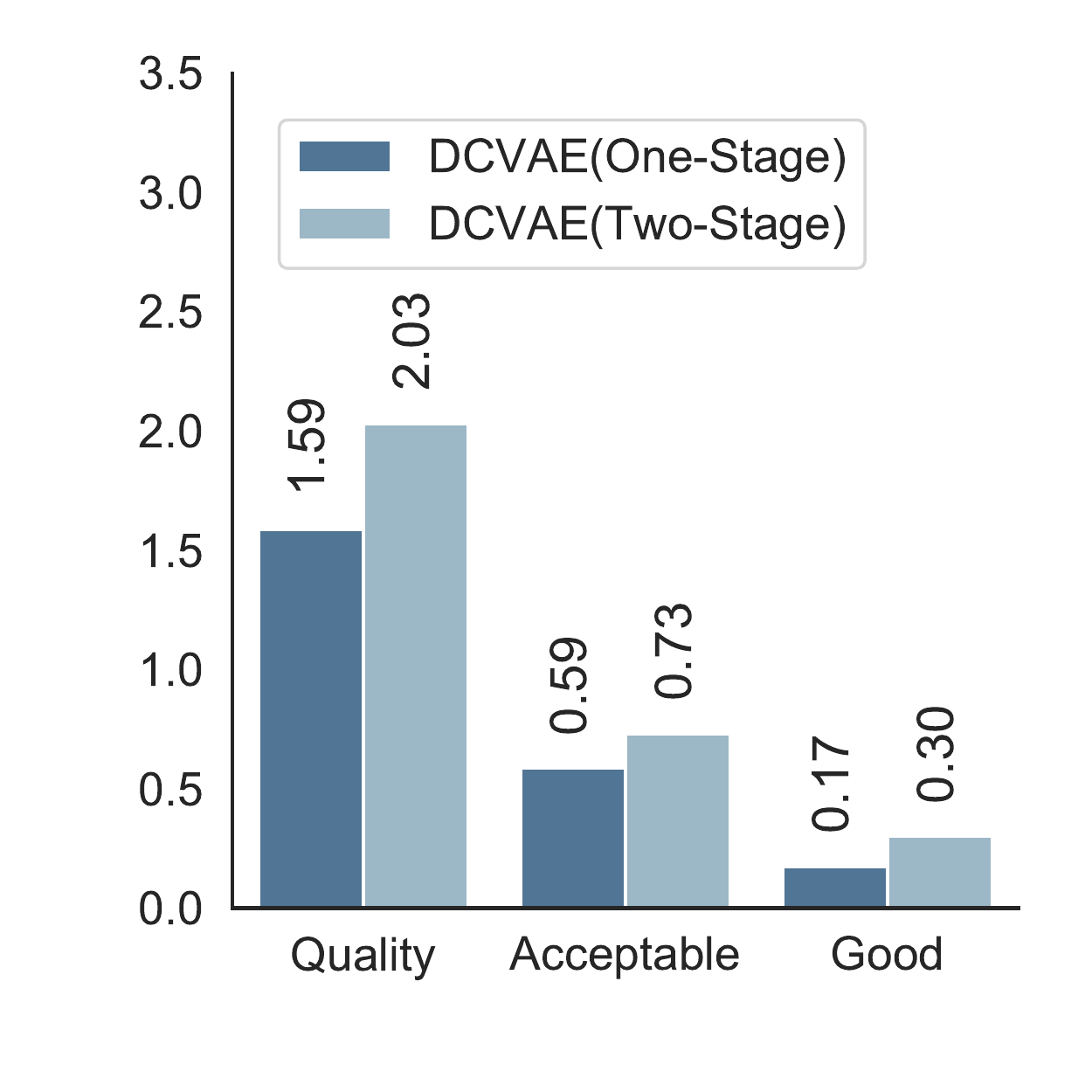}
 }
  \subfigure[]{\label{fig:dcvae_cd}
  \includegraphics[width=0.3\linewidth]{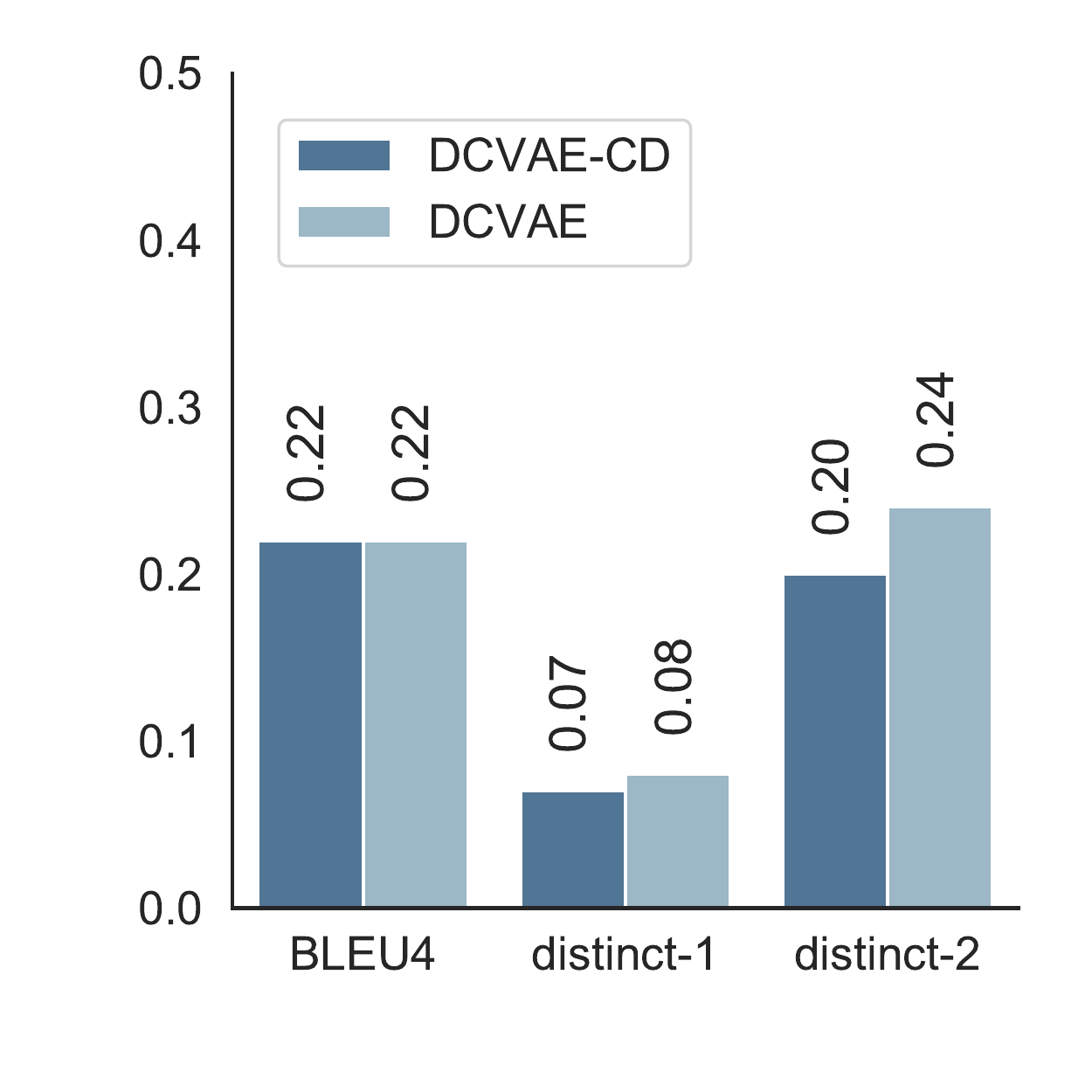}
 }
\end{center}
% \vspace{-5pt}
\caption{
(a)/(b): Automatic/human evaluation on the DCVAE with/without the two-stage sampling approach. 
%(b): Human evaluation on the DCVAE with/without the two-stage sampling approach. 
(c): Automatic evaluation on our proposed DCVAE and the modified DCVAE that uses a pure categorical distribution (DCVAE-CD) in which each variable has no exact meaning.}
\label{fig:two_stage}
% \vspace{-10pt}
\end{figure*}

\noindent
We further look into whether the two-stage sampling method is effective in the proposed DCVAE.
Here, the One-Stage method corresponds to the basic formulation in Section~\ref{sec:basic} with no use of the clustering information in the prior or posterior network.
Results on both automatic and human evaluation metrics are shown in Figure.~\ref{fig:stage_auto} and ~\ref{fig:stage_human}.
%
%To validate the effectiveness of our proposed two-stage sampling, we performed an evaluation with automatic metrics and human judgments used in our previous experiments. 
%We can observe that DCVAE with the two-stage sampling approach outperforms the method without it on both automatic and human evaluations. 
We can observe that the performance of the DCVAE without the two-stage sampling method drops drastically. 
%compares with the other compared methods.
%xxxx
This means that the proposed two-stage sampling method is important for the DCVAE to work well. 
%In terms of automated evaluation, the proposed two-stage sampling method achieves higher BLEU-4 scores and distinct ratio. 
%That means a model that uses a two-stage sampling method can generate more informative and diverse responses than those don't. Besides, the automated evaluation results are consistent with human judgments. From the results on the human evaluation, we can notice that the two-stage sampling method has better performance compared with the one-stage sampling method in terms of acceptable ratio, good ratio and diversity.

%\textcolor{red}{
Besides, to validate the effectiveness of clustering, we implemented a modified DCVAE (DCVAE-CD) that uses a pure categorical distribution in which each variable has no exact meaning. That is, the embedding of each latent variable does not correspond to any word embedding.  Automatic evaluation results of this modified model are shown in Figure.~\ref{fig:dcvae_cd}. We can see that DCVAE-CD performs worse, which means the distribution on word vocabulary is important in our model.

%%%%%%%%%%%%%%%%%%%%%%%%%%%%%%%%%%%%%%%%%%%%%%%%%%%%%%%%%%%%%%%%%%%%%%
\section{Conclusion}
%\vspace{-5pt}

In this paper, we have presented a novel response generation model for short-text conversation via a discrete CVAE. We replace the continuous latent variable in the standard CVAE by an interpretable discrete variable, which is set to a word in the vocabulary. The sampled latent word has an explicit semantic meaning, 
%which is able to capture short-text-level variations, 
acting as a guide to the generation of informative and diverse responses. We also propose to use a two-stage sampling approach to enable efficient selection of diverse variables from a large latent space, which is very essential for our model. % to work well. 
Experimental results show that our model outperforms various kinds of generation models under both automatic and human evaluations.
%, and generates more diverse and informative responses.
% \textcolor{red}{In current work, the model only use one keyword to generate a response.   We plan to extend the exploration to richer representation by sampling multiple keywords at the same time.
% And in the future, We also plan to extend the model to generate multiple responses simultaneously.}

\section*{Acknowledgements}
This work was supported by National Natural Science Foundation of China (Grant No. 61751206, 61876120).

\end{CJK*}
\bibliographystyle{acl_natbib}
\bibliography{ref}
\end{document}